\pdfoutput=1
\PassOptionsToPackage{table}{xcolor}
\documentclass{style}
\usepackage[utf8]{inputenc}
\usepackage{amsmath,amssymb,booktabs,graphicx,microtype,xcolor}
\usepackage{array,tabularx,enumitem,placeins,pdflscape,float}
\usepackage{tikz}
\usetikzlibrary{arrows.meta,positioning,calc,fit}
\usepackage{xurl}
\hypersetup{pdfsubject={Research preprint},pdftitle={RoboACE: Adversarial Co-Evolution for Robot Self-Improvement},pdfauthor={Delin Zhao, Zhengrong Yue, Junlin He, Yali Wang}}
\newcolumntype{Y}{>{\raggedright\arraybackslash}X}
\newcommand{\method}{RoboACE}
\newcommand{\initial}{\method{} (before evolution)}
\newcommand{\finalsys}{\method{} (after evolution)}

\newcommand{\piFive}{$\pi_{0.5}$}
\setlist{nosep,leftmargin=*}
\title{RoboACE: Adversarial Co-Evolution for Robot Self-Improvement}

\author{%
  Delin~Zhao\texorpdfstring{$^{*}$}{}\hspace{0.35cm}%
  Zhengrong~Yue\texorpdfstring{$^{*}$}{}\hspace{0.35cm}%
  Junlin~He{$^{*}$}{}\hspace{0.35cm}%
  Yali~Wang\texorpdfstring{$^{\dagger}$}{}%
}

\affiliation{%
  {\fontfamily{sansmedium}\fontsize{10.5}{13}\selectfont MMLab @ SIAT}
}

\contribution{\textsuperscript{*}Equal contribution\qquad\textsuperscript{\textdagger}Corresponding author}
\checkdata[\raisebox{-0.2em}{\includegraphics[width=0.025\linewidth]{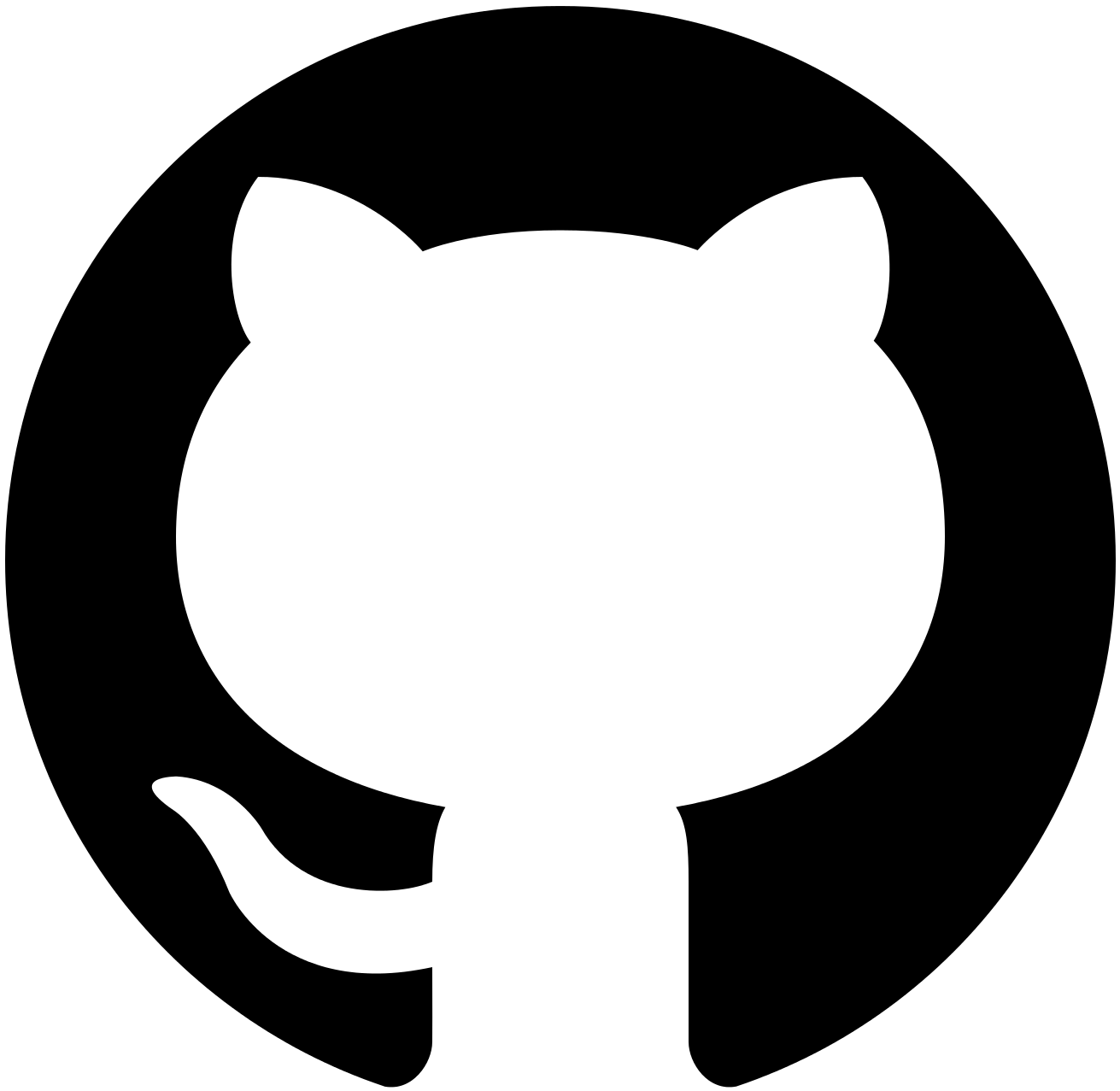}}~~GitHub Repo]{\href{https://github.com/DeLin1001/RoboACE}{\texttt{https://github.com/DeLin1001/RoboACE}}}

\abstract{
Robot agents can improve reusable skills, planning procedures, and recovery
strategies from execution feedback, yet a fixed practice source leaves the
search for new weaknesses unchanged. Useful practice depends on the agent's
current capabilities, so the procedure that chooses practice must also be
revisable. We present \method{} (Robot Adversarial Co-Evolution), a framework
that jointly evolves curriculum and execution harnesses around frozen
foundation models. The curriculum harness uses accumulated capability
evidence to propose informative challenges, while the execution harness
responds with reusable revisions that expand execution capabilities. Alternating updates hold one
side fixed as the other changes, and admitted challenges form fixed
snapshots for comparing execution revisions. Starting from a compact set
of LIBERO practice conditions, the resulting frozen deployment outperforms
the evaluated controllers and agent systems across four LIBERO-family
panels. On LIBERO-PRO, it reaches 74.0\% success, exceeding the initial
deployment by 15.0 percentage points and direct execution by its VLA
controller by 13.5 points. Case studies document changes to evidence-guided
practice selection, drawer-skill grounding, and controller intervention.
}

\begin{document}
\maketitle
\begingroup
\setlength{\intextsep}{4pt}
\captionsetup{font=footnotesize,skip=4pt}
\begin{figure}[H]
\centering
\includegraphics[width=.66\linewidth]{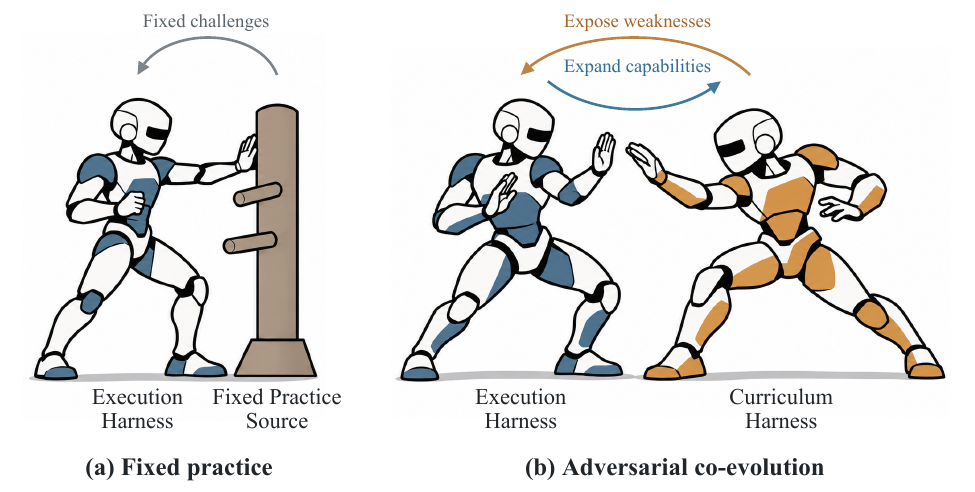}
\caption{\textbf{From fixed practice to adversarial co-evolution.}
(a) Fixed practice improves execution against an unchanged challenge source.
(b) In \method{}, the curriculum exposes current execution weaknesses,
while execution expands its capabilities through reusable harness revisions.
As capabilities grow, the curriculum targets new weaknesses,
sustaining adversarial co-evolution.
The blue and orange robots represent the execution and curriculum
harnesses, respectively.}
\label{fig:practice-partners}
\end{figure}
\endgroup
\clearpage

\section{Introduction}
Robot agents increasingly improve their own software through interaction:
RoboRSI refines reusable skills, Zetta develops critics and recovery skills,
and DynaHarness revises interfaces between semantic decisions and physical
execution
\citep{wen2026roborsi,ding2026zetta,deng2026dynaharness}.
Together with RAX, which turns exploration feedback into reusable programs,
these systems make execution procedures revisable \citep{huang2026rax}.
An equally important choice is what the agent practices next, because a
fixed source leaves unprobed weaknesses outside the feedback driving
revision. Improving execution alone leaves this choice unchanged.

Useful practice changes as an agent's capabilities develop. A drawer skill
that succeeds alone can still fail when composed with object placement,
so practice must probe the composition as well as its individual steps.
Selecting such practice requires interpreting prior outcomes, identifying
prerequisites, and retaining conditions that reveal regressions. Automatic
curricula already connect practice selection with skill accumulation
\citep{wang2023voyager}, but the selection procedure itself must also
adapt as failures are resolved and new interactions emerge. Earlier rules
for finding informative challenges can then become inadequate. This motivates a
robot self-improvement loop that revises how it chooses practice alongside
how it executes tasks.

We introduce \method{} (Robot Adversarial Co-Evolution), in which a
curriculum harness and an execution harness evolve through an adversarial
interaction (Figure~\ref{fig:practice-partners}). A harness contains the
prompts, skills, tools, interfaces, and agent-readable memory that organize
behavior around frozen foundation models. The curriculum side uses prior
capability evidence to expose weaknesses in the current execution
harness. The execution side responds with reusable revisions that expand execution capabilities, and
its changing capabilities redirect the curriculum's search. Independent
agents make runtime decisions on each side, while external optimizers
use the resulting evidence to install revisions for future runs.

Adversarial challenge generation requires a stable reference for judging
progress, so \method{} alternates the two updates. With execution
fixed, the curriculum side revises its procedures and proposes practice;
admission checks validity and consistency before fixing a comparison
snapshot. The execution side then evaluates candidate revisions on that
snapshot, and consolidated outcomes inform the next curriculum update.
This order separates challenge selection from the response it elicits,
while an infrequently invoked Strategic Advisor reviews longer histories
to reconsider repetitive or stalled revision directions.

\method{} brings together software self-improvement and evolving
curricula through persistent revisions to both harnesses. RPG constructs
practice from offline data and refines execution on the resulting fixed
task set \citep{wang2026rpg}; our curriculum side also revises the procedures
that select and construct subsequent practice. POET evolves environments
together with their solutions \citep{wang2019paired,wang2020enhanced};
our implementation instead adapts the software that organizes challenge
production and robot execution around frozen models. The changing task
library is an output of curriculum-harness evolution, alongside the
execution revisions it prompts.

We instantiate this loop in simulated manipulation, beginning with a compact
set of LIBERO practice conditions and evaluating frozen deployments on
LIBERO, LIBERO-Plus, LIBERO-PRO, and LIBERO-X. The final deployment outperforms
the evaluated systems on all four panels (Table~\ref{tab:main}).
On LIBERO-PRO, it improves over the initial deployment by 15.0 percentage
points and over direct execution by the same VLA controller by 13.5 points.
Case studies show how evidence-reading guidance changes independent
practice selection and how execution revisions clarify skill grounding
and controller handoffs. Together, the framework and empirical study
establish a concrete implementation of robot self-improvement in which
both the practice-selection procedure and the execution software remain
open to revision.

\section{Related Work}
\paragraph{Embodied agents around foundation models.}
Agent software connects foundation-model reasoning to physical control.
Code as Policies expresses closed-loop robot behavior as language-model
programs \citep{liang2023code}, while CaP-X evaluates robot coding agents
and the use of execution feedback for skill synthesis
\citep{fu2026capx}. RPent augments a controller with prompt-based experience
\citep{rpent2026harness}, and VoLo combines subgoal decomposition, VLM
monitoring, and grasp replanning around a VLA controller
\citep{volo2026agent}. Related systems also organize closed-loop control
through execution-aware agent harnesses \citep{du2026sera,lin2026fastharness}.
Our focus is the revision of such harnesses across rounds, together with
the procedures that determine which challenges those rounds contain.

\paragraph{Embodied self-improvement.}
Execution feedback can support persistent changes to robot software.
RoboRSI structures reusable skill improvement and provides the execution
infrastructure used in our implementation \citep{wen2026roborsi}.
Zetta develops runtime critics and recovery skills around a frozen policy
\citep{ding2026zetta}, and DynaHarness uses contracts and failure-driven
revisions to connect semantic decisions with execution
\citep{deng2026dynaharness}. RAX and ASPIRE convert exploration into
programs and skills \citep{huang2026rax,lu2026aspire}; EmbodiedEvo distills
physical feedback into code skills and diagnostic lessons
\citep{li2026embodiedevo}. RPG generates simulated practice from offline
data before refining prompts and skills on that task set
\citep{wang2026rpg}, while EmbodiedRSI selects experiments to distinguish
hypotheses about code and skill changes \citep{song2026embodiedrsi}.
\method{} extends the object of revision to the curriculum harness that
uses execution evidence to select and construct future practice.

\paragraph{Curriculum and challenge evolution.}
Changing challenges can guide learning toward abilities that a fixed task
source overlooks. Voyager connects automatic curricula with skill
accumulation \citep{wang2023voyager}, and Eureka generates reward functions
for robot training \citep{ma2024eureka}. POET jointly develops environments
and solutions \citep{wang2019paired,wang2020enhanced}, PAIRED uses regret
to guide environment design \citep{dennis2021emergent}, and asymmetric
self-play builds robotic curricula through proposer and solver roles
\citep{openai2021asymmetric}. Eurekaverse generates environment programs
\citep{liang2025eurekaverse}; OMNI-EPIC produces tasks and rewards in
response to learning progress \citep{faldor2025omniepic}; GenEnv co-evolves
agent and environment policies \citep{guo2025genenv}. In \method{}, both
the challenge-selection procedures and the execution procedures receive
persistent software revisions while foundation-model weights remain fixed.

\paragraph{Pretrained robot control.}
Generalist controllers supply action priors learned from large robot
datasets. OpenVLA and \piFive{} illustrate this approach
\citep{kim2025openvla,intelligence2025pi}, while OpenVLA-OFT improves the
fine-tuning and inference recipe and $\pi_0$-FAST studies efficient action
tokenization \citep{kim2025oft,pertsch2025fast}. Harness adaptation changes
how an agent invokes such controllers, combines them with other tools,
and responds to feedback. We use \piFive{} as the frozen controller
throughout development and revise the software around it.

\section{Adversarial Co-Evolution of Robot Harnesses}
\label{sec:method}

\begin{figure*}[t]
\centering
\includegraphics[width=\linewidth]{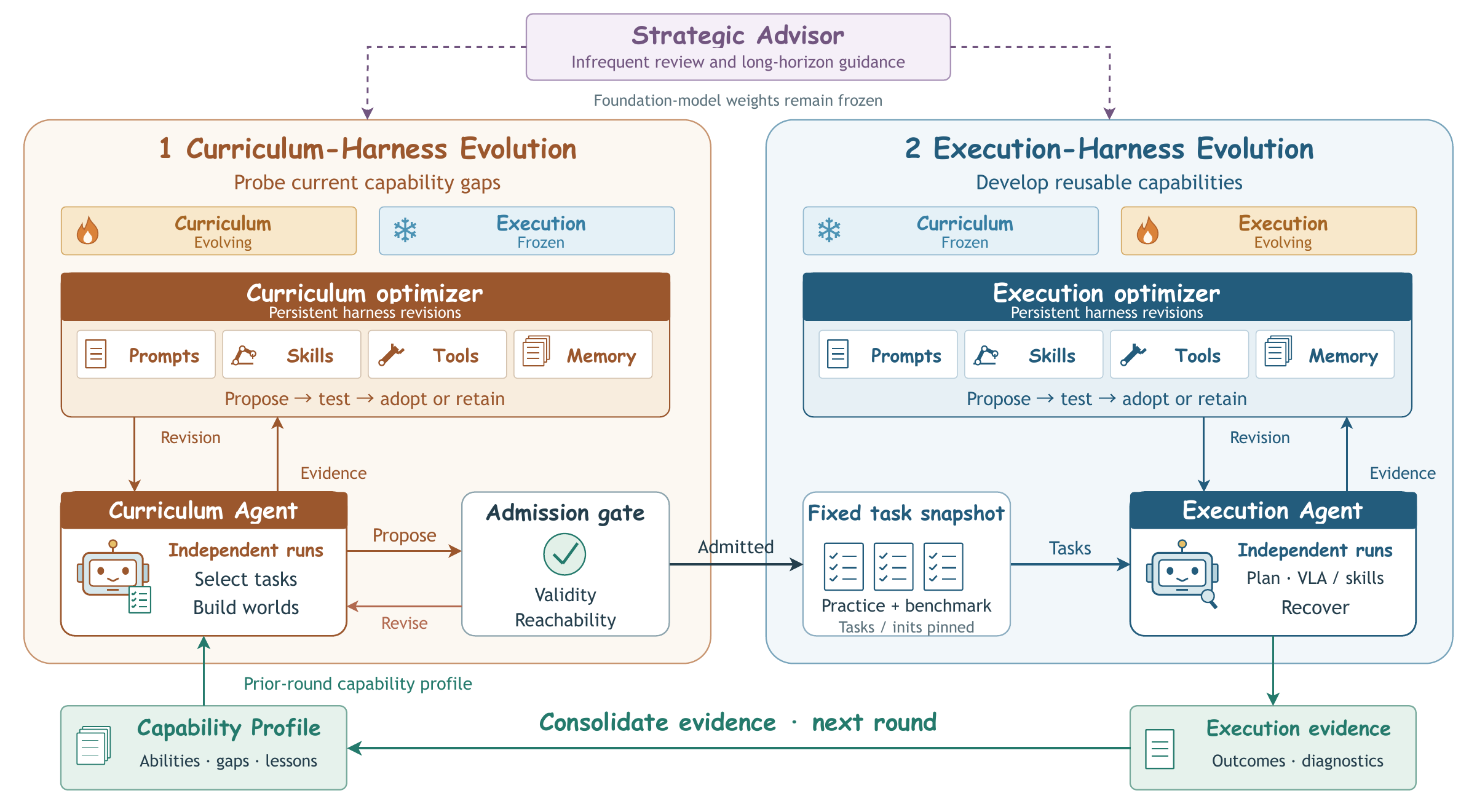}
\caption{\textbf{Alternating revisions connect challenge production and execution.}
The curriculum side uses prior capability evidence to probe a fixed
execution deployment. Admitted conditions form a fixed snapshot for
comparing execution revisions, whose outcomes inform the next round.
Independent agents run the harnesses, external optimizers revise them,
and the Strategic Advisor periodically reviews the longer history.}
\label{fig:framework}
\end{figure*}

\subsection{Framework Overview}
\label{sec:overview}
\method{} takes prior capability evidence and the current pair of harnesses
as input and returns revised harnesses together with evidence for the next
round. The curriculum harness guides practice selection and construction;
the execution harness organizes planning, code skills, and pretrained
control. Each contains reusable prompts, skills, tools, interfaces, and
agent-readable memory. All foundation-model weights remain frozen.
Figure~\ref{fig:framework} shows how the two sides exchange evidence while
updating in alternating phases.

Each side separates runtime decisions from persistent revision. An
independent \emph{agent} runs its harness to propose practice or attempt
a task. An external \emph{optimizer} examines the resulting evidence,
proposes changes, and tests them through further independent runs.
We call the accepted harness version \emph{Stable} and a proposed revision
\emph{Candidate}. A phase can retain its incoming Stable version, and its
revision history records the tests and the reasons for adoption or rejection.
The Strategic Advisor can review evidence across multiple local updates
without taking over these decisions.

In round $k$, the curriculum side receives the prior Capability Profile
$P_k$ and a fixed execution deployment. Its admitted proposals and the
internal benchmark form a fixed task snapshot $\mathcal T_k$. The execution
side first measures its incoming deployment on this snapshot, then tests
revisions and selects a deployment. Consolidated execution evidence yields
$P_{k+1}$. This ordering separates the curriculum's prediction of useful
practice from the difficulty and response measured afterward.
Appendix~\ref{app:implementation} specifies the operational protocol.

\subsection{Phase 1: Evolving the Curriculum Harness}
The curriculum must choose conditions that reveal actionable gaps in the
current execution capability. Its agent reads $P_k$, which summarizes
demonstrated abilities, applicable lessons, unresolved gaps, and uncertainty.
Using this profile, task statistics, and descriptions of available tasks
or assets, the agent retains, replaces, or adds practice conditions.
Selection balances new challenges with prerequisites and familiar
conditions that can expose regressions; increasing difficulty alone is
not the objective.

The curriculum optimizer revises how the agent makes these choices.
It tests whether a candidate harness supports better use of evidence,
maintains coverage, and recovers from failed retrieval or construction.
The revised assets persist across future curriculum-agent runs, whose
practice choices remain independent. Section~\ref{sec:cases} illustrates
this distinction through a revision to evidence-reading guidance and the
retrieval interface.

\paragraph{Admission and handoff.}
A challenge must qualify before it replaces an existing condition.
Native tasks retain their source definitions and initializations.
For constructed worlds, admission checks physical validity and reachability,
then a Judge checks agreement among the instruction, goal, and scene.
Failed or inconclusive checks return the proposal for revision.
These checks use no fresh VLA rollouts to pre-screen difficulty and do
not certify a complete executable solution. A separate Benchmark Curator
maintains reference coverage and core anchors. At handoff, the curriculum
deployment is fixed, and admitted practice and internal benchmark
conditions form $\mathcal T_k$. Task definitions and initializations
remain fixed throughout the execution phase.
Appendix~\ref{app:admission} details the checks.

\subsection{Phase 2: Evolving the Execution Harness}
The execution phase turns the admitted challenges into evidence for
reusable changes. The execution agent first attempts $\mathcal T_k$ with
the incoming Stable harness. It chooses among code skills and VLA control,
which can address a whole task or a shorter language goal.
The environment's native predicate determines task completion, so a
successful local operation, such as grasping an object, can still leave
the overall task unfinished.

The execution optimizer diagnoses failures from outcomes, trajectories,
public observations, and tool feedback. Diagnosis distinguishes failures
of a skill from misleading skill descriptions or inappropriate tool choices,
because these require different revisions. Candidate changes are tested
through independent agent runs with task definitions, initializations,
model settings, and budgets held fixed. Conditions shared by practice and
the internal benchmark use the same execution record.
Targeted runs diagnose individual mechanisms, while comparisons across
the full snapshot inform adoption. The optimizer considers successes,
regressions, reuse, and cost before installing Candidate or retaining Stable.

\subsection{Closing the Loop with Execution Evidence}
Execution outcomes must inform future practice without exposing evidence
reserved for other roles. The Experience Consolidator converts permitted
task-level results and lessons into $P_{k+1}$. The curriculum receives
this profile and task statistics, through which unresolved gaps can prompt
new challenges and resolved gaps can redirect practice. Raw images,
trajectories, model and tool calls, and implementation differences remain
on the execution side. Geometry reserved for the Judge remains outside
the execution agent's observations.

A diagnosis becomes useful to future agents only when it reaches a
deployable prompt, skill description, or retrievable manual. The optimizer
checks whether the agent actually receives and uses the revised material.
Persistent assets are versioned with each deployment and kept separate
from episode workspaces. Appendix~\ref{app:memory} describes this interface.
Final evaluation remains outside the loop and does not inform curriculum
choices or version selection.

\subsection{Long-Horizon Strategic Guidance}
Local revisions can repeatedly address nearby failures while leaving the
broader direction of improvement unexamined. The Strategic Advisor offers
an infrequent, on-demand review across multiple updates. It reads the
requesting optimizer's permitted evidence and revision history to identify
recurring failures, stalled revision patterns, and gaps in coverage.
Its output proposes directions for investigation rather than installing
harness changes.

The local optimizer translates this advice into experiments, evaluates
the evidence, and decides which revisions to adopt. Recommendations about
the inactive side wait for that side's next update phase. This separation
allows a longer view of the campaign while preserving the alternating
comparison protocol and local responsibility for adoption.

\section{Empirical Study}
\label{sec:setup}
We evaluate whether the final frozen deployment improves over its initial
version and alternative controllers or agent systems. We then examine
recorded revisions to identify how the curriculum and execution harnesses
changed during development.

\subsection{Development and Evaluation Scope}
We instantiated \method{} with GPT-6.1 (Terra) and pretrained tools including
SAM~3 and \piFive{} control \citep{intelligence2025pi}. The campaign
maintained 30 practice conditions and an internal benchmark of up to
40 conditions, with 27 saved round records. Both harnesses received
persistent software revisions; no LLM or VLA weights were trained or
fine-tuned by this project.

Development remained within the LIBERO ecosystem. The evolving practice
libraries included 76 distinct native LIBERO conditions across rounds,
together with constructed worlds that passed admission. Evaluation tests
transfer both to reserved LIBERO conditions and to LIBERO-Plus,
LIBERO-PRO, and LIBERO-X. Development uses only LIBERO conditions. The held-out LIBERO panel is condition-disjoint from the audited development history, while LIBERO-Plus, LIBERO-PRO, and LIBERO-X are not used during development.

We used four fixed panels of 200 conditions each
\citep{liu2023libero,fei2025libero,zhou2026liberopro,wang2026liberox}
(Table~\ref{tab:panels}). LIBERO contains 40 tasks with five reserved
initializations per task. LIBERO-Plus and LIBERO-PRO use fixed perturbation
subsets, while LIBERO-X includes 40 conditions at each challenge level,
including language-paired conditions. These panels define the evaluation
in this study and do not constitute the complete official benchmark
protocols.

\begin{table}[t]
\caption{\textbf{Four panels test transfer from the development campaign.}
Each contains 200 conditions. Development exposure refers to this project's
self-evolution campaign, not to pretrained-model training data.}
\label{tab:panels}
\centering\small
\renewcommand{\arraystretch}{1.18}
\begin{tabularx}{\linewidth}{@{}lYY@{}}
\toprule
{\bfseries Panel} & {\bfseries Coverage} & {\bfseries Evaluation role}\\
\midrule
LIBERO & 40 tasks, five reserved initializations & Transfer to reserved conditions\\
LIBERO-Plus & Seven perturbation categories & Transfer beyond development sources\\
LIBERO-PRO & Four available perturbation dimensions & Transfer beyond development sources\\
LIBERO-X & Five challenge levels & Transfer beyond development sources\\
\bottomrule
\end{tabularx}
\end{table}

\subsection{Compared Systems and Measurement}
The core comparison includes \initial{}, \finalsys{},
and \piFive{} \citep{intelligence2025pi}.
The \piFive{} baseline uses the final system's official VLA controller
continuously, without an upper-level LLM or other skills. Initial and
final agents use Terra/high at evaluation. The controller baseline shares
the final agent's controller, while the systems differ in orchestration
and available tools. We also evaluate released OpenVLA-OFT
\citep{kim2025oft}, $\pi_0$-FAST \citep{pertsch2025fast}, Cosmos Policy
\citep{kim2026cosmos}, and GR00T N1.7 \citep{nvidia2026gr00t17} without an
upper-level LLM.

The agent comparisons use RPent without target-benchmark experience
\citep{rpent2026harness} and VoLo \citep{volo2026agent}. RPent receives a
neutral prompt without target-experience retrieval. VoLo combines subgoal
decomposition, VLM monitoring, and grasp replanning. Both use Terra/high,
official \piFive{} control, and public observations without hidden
geometric state. Appendix~\ref{app:evaluation-details} specifies the
baseline variants, controller interfaces, and serving configurations.

Table~\ref{tab:main} includes model-author reports
\citep{openpi2025libero0,openpi2025libero05} and published benchmark
results \citep{fei2025libero}.

Our evaluation runs use the environment's native task predicate. All
systems in our evaluation share a per-episode budget of 8,000 environment
control steps at 20~Hz (6~min 40~s of simulation time).
\piFive{} executes continuous five-action chunks.

For our evaluations, each cell contains one designated observation for
every assigned condition. Historical prefix observations are retained without
choosing the best of repeated attempts. Success rate is the number of
confirmed native successes divided by all 200 conditions; timeouts,
infrastructure errors, and all other outcomes count as failures.
Only complete, independently audited panels are scored. The results
characterize one development trajectory and its resulting deployments.

\subsection{System-Level Performance}
The final \method{} deployment outperforms all evaluated systems on the
four fixed panels (Table~\ref{tab:main}). Relative to the initial
deployment, it improves by 35.5 percentage points on LIBERO and by
30.0 points on LIBERO-Plus. The gain is 15.0 points on LIBERO-PRO and
11.0 points on LIBERO-X. These comparisons measure the overall change
between the initial and final deployments, including their accumulated
harness revisions.

\begin{table*}[!t]
\caption{\textbf{System comparison across four LIBERO-family benchmarks.}
Success rates (\%), with best values in bold. Model baselines use checkpoints
fine-tuned on original LIBERO and evaluated without further training on
LIBERO-Plus, LIBERO-PRO, and LIBERO-X. \method{} evolves only on LIBERO
and is deployed frozen on held-out LIBERO and all three transfer panels.
The \piFive{} baseline runs the final \method{} controller alone.}
\label{tab:main}
\centering
\small
\setlength{\tabcolsep}{5pt}
\renewcommand{\arraystretch}{1.24}
\begin{tabularx}{\linewidth}{@{}Y*{4}{>{\centering\arraybackslash}p{.15\linewidth}}@{}}
\toprule
{\bfseries System} & {\bfseries LIBERO} & {\bfseries LIBERO-Plus} & {\bfseries LIBERO-PRO} & {\bfseries LIBERO-X}\\
\midrule
\piFive{} \citep{intelligence2025pi} & 96.9 & 92.0 & 60.5 & 11.5\\
OpenVLA-OFT \citep{kim2025oft} & 96.8 & 75.5 & 51.0 & 2.5\\
$\pi_0$-FAST \citep{pertsch2025fast} & 85.5 & 61.6 & 49.0 & 4.5\\
Cosmos Policy \citep{kim2026cosmos} & 98.5 & 88.5 & 52.5 & 2.5\\
GR00T N1.7 \citep{nvidia2026gr00t17} & 97.5 & 89.0 & 48.0 & 2.5\\
\midrule
RPent \citep{rpent2026harness} & 92.0 & 84.0 & 32.5 & 12.5\\
VoLo \citep{volo2026agent} & 97.5 & 90.0 & 60.5 & 10.0\\
\midrule
\initial{} & 63.5 & 64.5 & 59.0 & 19.5\\
\rowcolor{pkured!7}
{\bfseries \finalsys{}} & {\bfseries 99.0} & {\bfseries 94.5} & {\bfseries 74.0} & {\bfseries 30.5}\\
\bottomrule
\end{tabularx}
\par\vspace{3pt}
{\footnotesize\raggedright RPent uses no target-benchmark experience.\par}
\end{table*}

\FloatBarrier
\suppressfloats[t]
\begin{figure*}[t]
\centering\includegraphics[width=\linewidth]{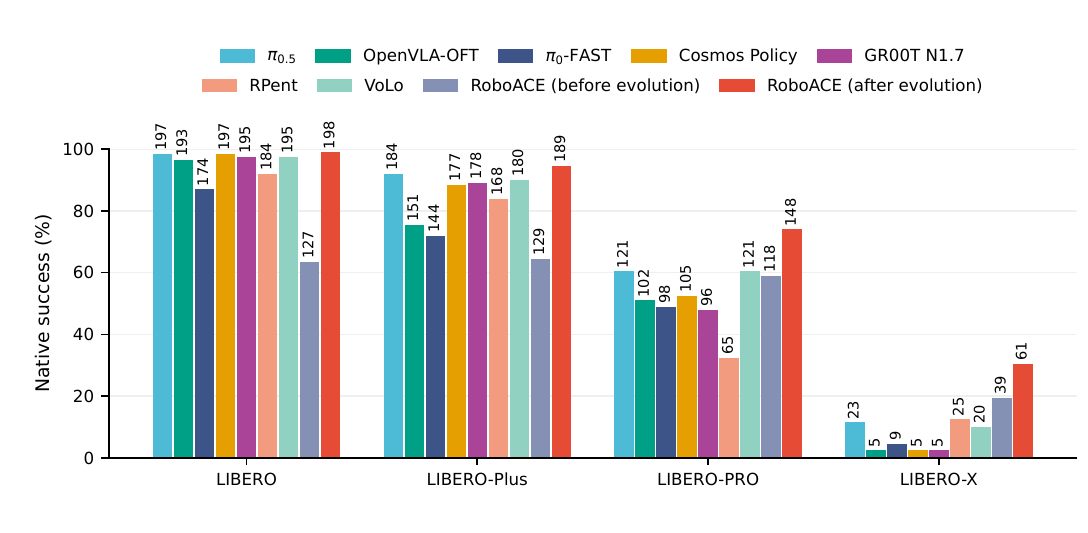}
\caption{\textbf{The final deployment improves across all four evaluation panels.}
Bar labels give success counts on a 200-episode scale. Model baselines use
checkpoints fine-tuned on original LIBERO and evaluated without further
training on LIBERO-Plus, LIBERO-PRO, and LIBERO-X.}
\label{fig:baseline-comparison}
\end{figure*}

\piFive{} is close to the final deployment on LIBERO and LIBERO-Plus,
while the separation grows on LIBERO-PRO and LIBERO-X
(Figure~\ref{fig:baseline-comparison}). On LIBERO-PRO, \method{} exceeds
\piFive{} by 13.5 percentage points; on LIBERO-X, the margin is
19.0 points. Sharing the official controller makes this comparison
informative about the complete agent system built around pretrained
control. 
\method{} also exceeds the strongest evaluated agent baseline on every
panel. Its margins are 1.5 and 4.5 percentage points on LIBERO and
LIBERO-Plus, respectively, and 13.5 and 18.0 points on LIBERO-PRO and
LIBERO-X.

\subsection{Case Studies of Harness Evolution}
\label{sec:cases}
\begin{figure*}[!t]
\centering
\includegraphics[width=0.7\linewidth]{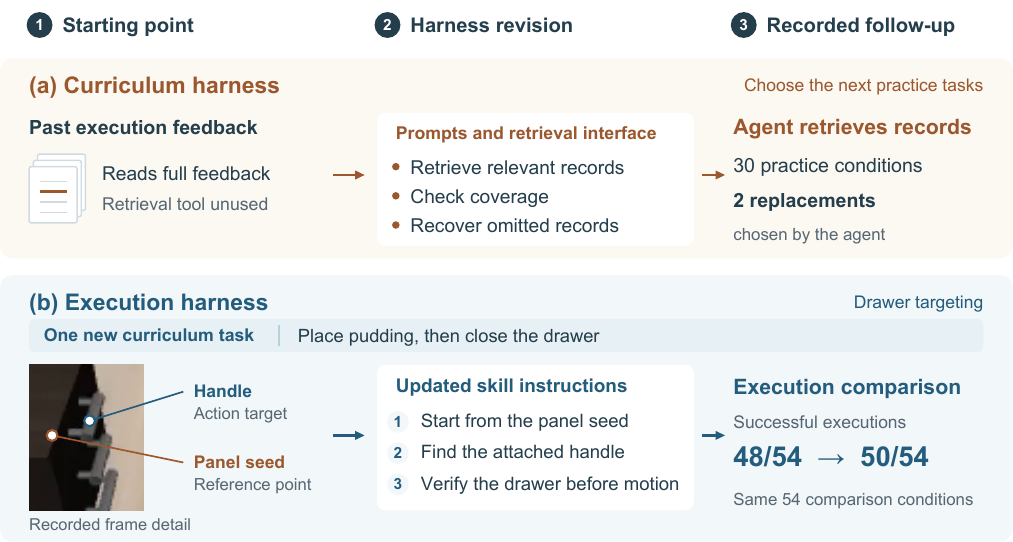}
\caption{\textbf{Reusable revisions address concrete gaps in both harnesses.}
Each row follows a case from its starting point to a reusable revision
and the recorded follow-up. (a) The curriculum revision changes how the agent
retrieves past execution feedback to select practice. A subsequent
independent run uses the reader to create a valid 30-condition course
with two agent-selected replacements. (b) Following new drawer practice
selected by the curriculum, the execution revision distinguishes a front-panel reference
point from the attached handle and requires target-drawer verification.
Successes change from 48 to 50 on the same 54-condition snapshot. The
frame is recorded, and the markers and rule summaries are explanatory.}
\label{fig:harness-revisions}
\end{figure*}

\paragraph{Curriculum revision changes how evidence is used.}
An optional compact evidence reader initially went unused. A revision to
prompts and the retrieval interface explained selective reading, coverage
checks, and targeted recovery after truncation
(Figure~\ref{fig:harness-revisions}(a)). In a subsequent independent run,
the curriculum agent used the reader and produced a valid 30-condition
course with two replacements of its own choosing. The persistent revision
concerned the procedure for using evidence, while the agent retained
responsibility for selecting the practice conditions.

\paragraph{Execution revision clarifies a skill's grounding contract.}
Earlier evidence indicated that standalone drawer skills did not reliably
compose. Before receiving fresh execution outcomes, the curriculum agent
retained 27 conditions and added top-drawer closure, pudding placement
followed by closure, and bottom-drawer opening. The ensuing execution
revision clarified the difference between a front-panel localization seed
and the attached handle (Figure~\ref{fig:harness-revisions}(b)). Skill
descriptions now required resolving that handle and verifying the target
drawer before motion. Successes across the full 54-condition comparison
snapshot changed from 48 to 50. The record connects a curriculum-selected composition challenge to a
persistent clarification of the execution interface.

\begin{figure*}[!t]
\centering
\includegraphics[width=0.7\linewidth]{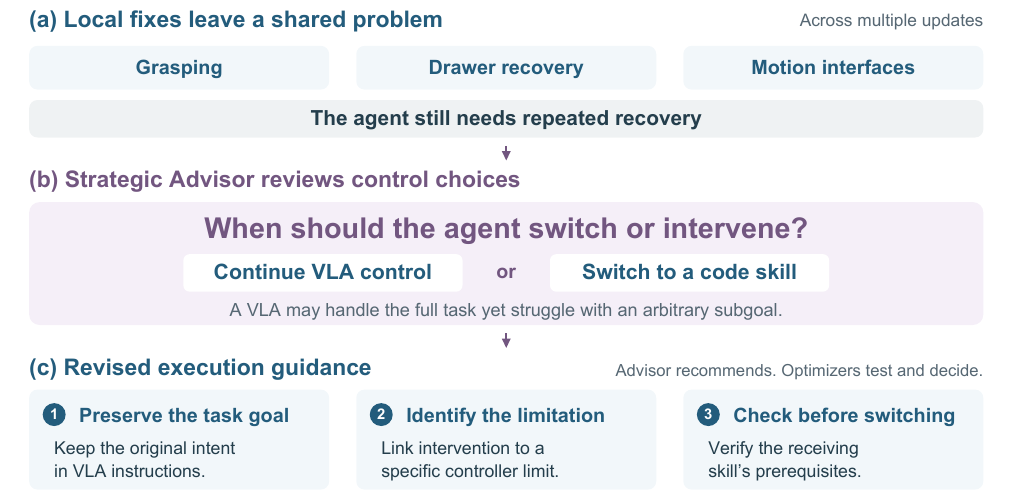}
\caption{\textbf{Strategic review turns recurring failures into broader execution guidance.}
(a) Repeated recovery remains a shared bottleneck across local revisions
to grasping, drawer recovery, and motion interfaces. (b) Longer-horizon
review asks when to continue VLA control, switch to code skills, or
intervene; whole-task capability does not guarantee reliable behavior
under arbitrary subgoals. (c) Revised guidance preserves task intent,
targets specific limitations, and checks handoff prerequisites. The
Advisor proposes directions; optimizers test revisions and decide adoption.}
\label{fig:advisor-case}
\end{figure*}

\paragraph{Strategic review connects failures across local revisions.}
Repeated recovery remained a shared bottleneck after changes to grasping,
drawer recovery, and motion interfaces. The Strategic Advisor reviewed
this history and identified a broader question: when should the agent
continue VLA control, switch to a code skill, or intervene in an ongoing
attempt (Figure~\ref{fig:advisor-case})? Whole-task VLA capability does
not ensure reliable behavior under arbitrary language subgoals.
The resulting guidance preserved task intent, targeted specific controller
limitations, and required checking prerequisites before a handoff.
The record illustrates how longer-horizon review can connect recurring
local failures to a common question about control selection.

\section{Conclusion}
We presented \method{}, a robot self-improvement framework that revises
both the selection of practice and the software used to execute it.
A curriculum harness proposes challenges from prior capability evidence,
and an execution harness responds with persistent revisions around
frozen foundation models. Alternating phases connect the two through
admitted tasks, fixed comparison snapshots, and consolidated feedback.

The final deployment achieves the highest success rate among the evaluated
systems on four LIBERO-family panels, with gains of up to 35.5 percentage
points over the initial deployment. Recorded cases show changes to
practice selection, skill grounding, and controller intervention.
These results support adversarial harness co-evolution as a concrete
approach to joint curriculum and execution revision. Establishing its
behavior across independent campaigns, broader environments, and physical
robots is the next empirical step.

\raggedbottom
\bibliographystyle{plainnat}
\bibliography{references}
\appendix
\section{Implementation and Evaluation Details}
\label{app:implementation}
This appendix specifies the update boundaries, task admission, deployable
memory, and baseline evaluation underlying
Sections~\ref{sec:method} and~\ref{sec:setup}.

\subsection{Operational Update Protocol}
\begin{table}[H]
\caption{\textbf{Alternation fixes the reference during each revision phase.}
Either phase can retain its incoming Stable version. Final evaluation
evidence remains outside the loop.}
\label{tab:protocol}
\centering\small
\renewcommand{\arraystretch}{1.18}
\begin{tabularx}{\linewidth}{@{}p{.11\linewidth}YY@{}}
\toprule
{\bfseries Step} & {\bfseries Independent work and revision} & {\bfseries Fixed reference or output}\\
\midrule
1 & Curriculum agent proposes practice; its optimizer tests reusable revisions. & Execution deployment and prior profile $P_k$.\\
2 & Validate proposals, retain or replace practice, and curate reference coverage. & Fixed comparison snapshot $\mathcal T_k$.\\
3 & Evaluate the incoming execution deployment and compare reusable revisions. & Same tasks, initializations, settings, and budgets.\\
4 & Adopt Candidate or retain Stable; consolidate permitted evidence. & Selected deployment and next profile $P_{k+1}$.\\
\bottomrule
\end{tabularx}
\end{table}

The curriculum proposes new conditions using prior capability evidence;
fresh execution observations arrive only after handoff. The curriculum phase uses only the prior-round capability profile; execution outcomes on newly admitted conditions are collected after the task snapshot is fixed. Candidate execution comparisons preserve
task definitions and initializations. These boundaries distinguish practice
selection, incoming difficulty, candidate behavior, and the evidence used
for adoption (Table~\ref{tab:protocol}).

\subsection{Task Validation and Admission}
\label{app:admission}
Native tasks preserve their source definitions and initialization identities.
Constructed tasks additionally pass compilation and physical-validity
checks for loadability, stable finite states, valid goal references, and
absence of severe interpenetration. Geometry checks use the compiled robot
and inverse kinematics to assess the reachability of key interaction
endpoints. The Judge then reviews semantic agreement among the instruction,
goal, and actual scene.

Failed or inconclusive proposals return to the curriculum agent for
revision. Constructed tasks are admitted after checks for compilation, physical validity, endpoint reachability, and instruction-goal-scene consistency. These checks do not guarantee full dynamic executability, and task difficulty is not pre-screened with fresh VLA rollouts. Recorded selections include both
native tasks and admitted constructed worlds.

\subsection{Deployable Memory Interface}
\label{app:memory}
The implementation exposes up to 2500 characters from each skill's main
description and supports retrieval of additional task and skill manuals.
The optimizer checks whether the agent receives and uses the relevant
material. An optimizer journal can contain a useful diagnosis without
making it accessible to future agents; deployment requires incorporating
it into an exposed description, prompt, or retrievable manual.
Episode workspaces remain separate from persistent assets, whose code,
tool, and memory versions are fixed with the deployment.

\subsection{Evaluation Identities and Controller Interfaces}
\label{app:evaluation-details}
Direct \piFive{} continuously uses the final system's official controller,
executing five actions per prediction chunk without an upper-level LLM
or other skills. Initial and final agents use Terra/high.
All systems in our evaluation share a per-episode budget of 8,000
environment control steps at 20~Hz (6~min 40~s of simulation time).

\paragraph{Released-model baselines.}
OpenVLA-OFT uses the released combined spatial/object/goal/10 fine-tuned
checkpoint with native eight-action chunks, front and wrist images,
and proprioception. $\pi_0$-FAST uses the official
\path{pi0_fast_libero} checkpoint, released image, state, and action
transforms, and five-action replanning from a ten-action horizon.
This project neither trains nor fine-tunes these models.

\paragraph{Agent-baseline variants.}
RPent uses a neutral prompt without target-experience retrieval, rather
than its experience-augmented variant. VoLo retains its deployed adapter's
subgoal decomposition, monitoring, and replanning, including VLM monitoring
and \texttt{replan\_grasp}. Both use Terra/high, official \piFive{} control,
and public observations without hidden geometric state.

\subsection{Complete-Panel Checks and Recording}
A released-model panel requires 200 unique request/result pairs, settled
workers, and an independent audit of source identity, raw records, and
outcome counts. Agent-baseline panels likewise require 200 complete,
independently audited conditions. Partial panels are excluded from scored
results.

Each condition contributes one designated main-table observation.
Historical prefix observations remain in place without best-attempt
selection. Timeouts and infrastructure outcomes remain in the denominator
and count as failures. Separate reassessments and targeted diagnostics do
not replace the designated main-table observations.

\end{document}